\documentclass[runningheads]{llncs}
\usepackage[T1]{fontenc}
\usepackage{graphicx}
\usepackage{framed} 
\usepackage{url} 
\usepackage{xcolor} 
\usepackage{tipa}
\begin{document}
\title{Value-Aware Multiagent Systems}
%
%
\author{Nardine Osman\orcidID{0000-0002-2766-3475}}
\authorrunning{N. Osman}
%
\institute{Artificial Intelligence Research Institute (IIIA-CSIC), Barcelona, Catalonia, Spain 
\email{nardine@iiia.csic.es}}
\maketitle              
\begin{abstract}
This paper introduces the concept of value awareness in AI, which goes beyond the traditional value-alignment problem. Our definition of value awareness presents us with a concise and simplified roadmap for engineering value-aware AI. The roadmap is structured around three core pillars: (1) learning and representing human values using formal semantics, (2) ensuring the value alignment of both individual agents and multiagent systems, and (3) providing value-based explainability on behaviour. 
The paper presents a selection of our ongoing work on some of these topics, along with applications to real-life domains.

\keywords{Value awareness \and Value alignment \and Value learning \and Value representation \and Real-life applications}
\end{abstract}
\section{Value Awareness}
There is a pressing need to ensure that the AI systems that we build are not only ethical and beneficial, but also align with our human values. Stuart Russell argues that we should change the goals of the field of AI itself; ``instead of pure intelligence, we need to build intelligence that is provably aligned with human values”~\cite{russell2019human}. This is now known as the value-alignment problem; that is, how to develop systems whose behaviour is aligned with human values. 

When studying the alignment of individual agents, the tendency is to reason about the individuals' decision-making processes~\cite{TostoD12,ijcai2017p26}. While studying the alignment of multiagent systems has led to reasoning about the norms of that multiagent system~\cite{abs-2110-09240,MontesS21,SerramiaLR20}, as norms have been the traditional means for mediating the behaviour of collectives (organisations, communities, or simple aggregates of individuals).

In this paper, we go beyond the traditional value alignment problem and introduce the notion of value awareness. We define value awareness as follows.

\begin{framed}
\noindent
\textcolor{gray}{\large \bf value-aware AI}\\
\textcolor{gray}{Noun [U]}\\
\textcolor{gray}{\textipa{/"v\ae{}l.ju\textlengthmark \tipathickspace \textschwa"we\textschwa{}r  ""e\i"a\i/}}

\noindent an AI system that identifies and understands a human's value system, abides by that value system, and explains its own behaviour and that of others in terms of that value system
\end{framed}

This definition lays the roadmap for value engineering research, an emergent field in AI dedicated to the engineering of value-aware systems~\cite{vale2023,vale2024}. First, to {\bf identify and understand} human value systems, the AI should be capable of \emph{learning} relevant values, and \emph{modelling} those values through formal semantics. Of course, value systems may exist on the individual level and the collective level. As such, the AI should also be capable of \emph{aggregating} individual value systems into one that represents the collective. Alternatively, it should be capable of \emph{using agreement mechanisms}, such as argumentation and negotiation, to help the collective agree on their value system. 

Second, to {\bf abide} by a value system, \emph{value-alignment} mechanisms are needed for both agents and multiagent systems, such as developing value-aligned decision making and value-aligned norm synthesis. The objective is to ensure behaviour of agents and multiagent systems is aligned with relevant values. 

Third, to {\bf explain} one's own behaviour or that of others in terms of value systems, then there is a need for developing \emph{value-based explainability} mechanisms.
     
We note that value-awareness research is not limited to raising awareness on the agent and multi-agent level, but also raising awareness for humans so they better understand their behaviour. As we see in Section~\ref{sec:aapplications}, some of the work in real-life application domains focuses on ensuring humans, like medical professionals or firefighters, better understand which values their (potential) actions are promoting. In other words, the AI also supports humans to make value-informed decisions. 

The remainder of this paper provides an overview of a selection of our ongoing work on various topics of this concise and simplified roadmap, along with applications to real-life domains.

\section{Selected Contributions}

\subsection{On Identifying and Understanding Human Values}\label{sec:valueLearning}
\subsubsection{Value Representation}
The first challenge is that of value representation. While work on values in AI is gaining ground, there is no formal model yet for the representation of human values. 
Furthermore, values in existing AI literature have usually been specified through labels (such as ‘fairness’), without any formal specification of the semantics of those values. 

In~\cite{aamas2024}, we propose a value-based taxonomy for the modelling of human values. This allows values to be organised hierarchically, where abstract concepts branch into concrete concepts. Property-based leaf nodes allow us to formally specify value semantics, which enables computational reasoning about values. This is because these property nodes essentially define how a value may be interpreted and assessed. The taxonomy also allows us to explicitly specify value relations and value importance, all of which are crucial elements for deliberating and reasoning over values. Furthermore, we illustrate how the proposed value-based taxonomy is aligned with the values literature in social psychology~\cite{arxiv2024}.

\subsubsection{Value Learning}
One of the main challenges of value awareness engineering is identifying and understanding relevant human values. It is not straightforward for stakeholders to identify and articulate their relevant values. Values are abstract constructs whose exact meaning (semantics) may change over time and context, and even from one person to another. Existing literature illustrates how numerous definitions exist today for values such as fairness or equality. Income inequality alone has been formalised through numerous equations, such as the Gini Index~\cite{ceriani2012}, Palma Ratio~\cite{cobham2013}, Theil Index~\cite{conceiccao2000}, and many many more~\cite{de2007}. 

We argue that identifying the semantics of values is context dependent, and should involve learning what those values mean for the relevant stakeholders. The learning process should take into account human feedback, which could either be obtained through dedicated user studies or by simply observing the human's behaviour. Our ongoing work with Hospital del Mar, Barcelona aims at identifying the formulae that best describe the semantics of the four basic bio-ethical values (or principles): beneficence, non-maleficence, autonomy and justice~\cite{Beauchamp2019}. In this work~\cite{valawai24a}, we are compiling a corpus of patient cases where each case is defined by a set of criteria that specify the patient's state, an action that is performed on that patient, and the change in the patient's state after the action is taken. An ongoing user study aims at having doctors annotate these patient cases with information on which of the four values is being promoted, demoted, or not affected. The results of this user study will then be used by an evolutionary strategy algorithm that aims at navigating the space of potential formulae to learn the formulae that best fits the data, and hence, best describes the doctors' view of how to understand the alignment of each of those values.  


\subsection{On Abiding by Human Values}
Our initial work on the alignment of multiagent systems with human values has mostly focused on the alignment of norms, as norms are what govern the behaviour of these systems. In~\cite{abs-2110-09240}, a basic formal approach for value alignment is presented, where the alignment of a given norm with a given value is assessed by the level of promotion of that value in future states of the world. Norms are understood as changing the future states of the world, and values (such as gender equality) are defined through equations that specify how states of the world may be assessed with respect to that value (such as checking the gender pay gap).  

\cite{MontesS21} builds on this initial work to present tools for norm synthesis that would optimise for certain values, tools that use the Shapley value concept from game theory to help assess the contribution of the different norms towards promoting those values, as well as tools for checking the compatibility of values under certain norms. The proposed work is tested in a taxing game, where the norms that best promote values like fairness and equality are synthesised and assessed.

\cite{valawai23b} proposes an approach that empowers agents by using theory of mind to reason about each others' values. The proposed mechanism allows agents to analyse norms not only from the perspective of their own value system, but from the perspective of other agents' value systems. This could eventually help agents when negotiating over the norms that best suit their collective. 

Concerning the behaviour of individual agents and their decision making process, we propose in~\cite{valawai24b} an approach for enhancing automated negotiation mechanisms with social values to help agents reach agreements by considering not only their individual utilities, but also social values such as fairness and equality. 


%

\subsection{Real-World Applications}\label{sec:aapplications}
In~\cite{alba24}, agent-based simulation is being used to analyse norms (policies) from the perspective of the values of fighting inequality and discrimination. Policies are categorised as aporophobic\footnote{The term aporophobia was coined by the philosopher Adela Cortina to describe having feelings of fear and an attitude of rejection of the poor~\cite{cortina2022}.} and non-aporophobic by experts in the field, and agent-based simulation is used to better assess the impact of aporophobia on inequality. Results show that aporophobic policies do in fact lead to larger inequality, compared to non-aporophobic ones.

In the medical field, we are working closely with Hospital del Mar, Barcelona to develop a system that could provide feedback to medical professionals on the alignment of their potential actions with the four basic bioethical values (or principles)~\cite{icaart24}. Our proposal is based on analysing the potential outcomes of an action. Alignment is then computed based on the value semantics learnt (see Section~\ref{sec:valueLearning}). For example, if the value ``comfort'' was of utmost importance, then actions that lead to states of the world where the patient suffers will not be considered aligned with that value. A multi-objective Markov decision process (MOMDP) is then used to help assess the alignment of entire medical protocols. 

A similar approach to our work in the medical field is also being used to help train firefighters. While fire departments have clear and well defined values, it is common to see these values differ from one geographical location to another. Furthermore, new firefighter students usually join with their own value systems, and training over values in agent-based simulations could help them become more aligned with their fire department's value system.

\section{Conclusion}
This paper has introduced the notion of value awareness in AI, and presented a concise and simplified roadmap for the development and engineering of value aware AI. It also listed some selected ongoing works covering various research challenges, along with a number of real-life applications. 

While the topic of values in AI is becoming more prominent~\cite{vale2023,vale2024}, the open challenges are numerous, and the proposed roadmap provides only a glimpse into what future research can delve into. Some of the presented challenges are just starting to get traction, such as the work on aggregating individual value systems into a value system for the collective~\cite{Lera-LeriBSLR22}. 
It is also evident that research on value-based explainability remains underdeveloped. To address this gap, we intend to build on our symbolic approach for value representation and value-based reasoning, which could provide the foundations for value-based explainability.

\begin{credits}
\subsubsection{\ackname} This work has been supported by the EU-funded VALAWAI (\#~101070930) project and the Spanish-funded VAE (\#~TED2021-131295B-C31) and Rhymas (\#~PID2020-113594RB-100) projects. 

\end{credits}
%
%
%
\bibliographystyle{splncs04}
\bibliography{mybibliography}

\end{document}